# CKmeans and FCKmeans : Two deterministic initialization procedures for Kmeans algorithm using a modified crowding distance

Abdesslem Layeb

LISIA Laboratory, NTIC faculty, university of Constantine2

Abdesslem.layeb@univ-constantine2.dz

**Abstract:** This paper presents two novel deterministic initialization procedures for K-means clustering based on a modified crowding distance. The procedures, named CKmeans and FCKmeans, use more crowded points as initial centroids. Experimental studies on multiple datasets demonstrate that the proposed approach outperforms Kmeans and Kmeans++ in terms of clustering accuracy. The effectiveness of CKmeans and FCKmeans is attributed to their ability to select better initial centroids based on the modified crowding distance. Overall, the proposed approach provides a promising alternative for improving Kmeans clustering.

1. Introduction

K-means is a popular clustering algorithm that divides a dataset into K clusters based on the similarity of data points. The algorithm is efficient and easy to implement but has some limitations, such as sensitivity to initial cluster centers, local optima, and assumptions about cluster shapes. The algorithm proceeds by initializing K centroids, assigning data points to the nearest centroid, computing the mean of the data points in each cluster, and repeating until convergence [1,2].

Despite its advantages, K-means clustering has some important considerations. The initial placement of centroids can affect the final clustering result, so multiple runs with different initializations may be necessary. The choice of "k" is also crucial and can be determined by domain knowledge or using techniques such as silhouette analysis or Elbow method. K-means may not always converge to the global optimal solution and is sensitive to outliers, which can lead to biased clustering results. Scaling of features is also necessary to avoid biased clustering results.

The initialization of K-means is a well-known challenge in the clustering community. In this paper, we address this challenge by proposing two novel initialization procedures for K-means: CKmeans and FCKmeans. The CKmeans procedure uses a modified crowding distance approach, inspired by the multi-objective optimization literature [3], to select the most representative initial centroids. The FCKmeans procedure, on the other hand, selects the furthest crowded points as initial centroids. Both procedures aim to provide a good deterministic initialization procedure that can improve the clustering results of K-means.

CKmeans and FCKmeans are designed to address the limitations of standard initialization methods, such as the sensitivity to the initial placement of centroids and the assumption of spherical clusters with equal variance. The experimental results presented in Section 4 demonstrate the superiority of our proposed procedures over standard initialization methods, such as K-means++ and random initialization, in terms of clustering accuracy and stability.

Our proposed procedures have potential applications in various fields, such as image and text clustering, where the performance of K-means heavily depends on the initialization of centroids. Section 2 provides an overview of existing kmeans initialization procedures. Section 3 presents a detailed description of the proposed initialization procedures and their implementation.

## 2. Initialization kmeans methods

K-means is a popular clustering algorithm used in machine learning and data analysis. The algorithm requires the initialization of K centroids before it can partition the data points into K clusters. The quality of the initialization can significantly affect the performance and final clustering results of the algorithm. In this section, we will discuss several initialization methods for K-means clustering algorithm.

One of the most popular initialization methods is k-means++, which was proposed by Arthur and Vassilvitskii in 2007 [4]. The k-means++ algorithm selects the first centroid randomly from the data points, and then selects the next centroids with a probability proportional to the distance from the data point to the nearest existing centroid. This approach tends to choose centroids that are well-spaced and can lead to better clustering results.

Another initialization method is the Maxmin method [5], which selects the first centroid randomly from the data points, and then selects the subsequent centroids by choosing the data point with the maximum distance to the nearest previously selected centroid. This approach can also lead to well-spaced centroids, but it can be sensitive to outliers.

The PCA-Part method [6] starts with a single cluster containing all data points and completes the process in k-1 steps. In each step, it selects the cluster with the largest partial sum of squared Euclidean distances and divides it into two separate sub-clusters using a hyperplane that passes through the cluster centroid and is orthogonal to the direction of the principal eigenvector of the covariance matrix.

The global k-means (GKM) [7] method starts with two centers and adds a center at each time by considering each data point as a candidate for the next center. The data point that leads to the minimum value of the objective function, which is the sum of squared Euclidean distances to the closest center, is selected as the next center. The modified GKM (MGKM) method proposes a different way to minimize an auxiliary cluster function to select the starting point of the next center. The fast MGKM (FMGKM) method exploits information gathered in previous iterations of the incremental algorithm to decrease memory usage and increase speed.

To tackle initialization problems of the k-means algorithm, the MinMax [8] version of k-means changes the objective function and uses maximum intra-cluster variance ($\varepsilon_{max}$) as a potential objective function to be minimized. The MinMax assigns a weight for each cluster, such that clusters with larger intra-cluster variance are allocated higher weights, and these weights are learned automatically. This approach is less affected by initialization and can discover high-quality solutions, even with bad initial centers.

## 3. CKmeans and FCkmeans presentation
### 3.1 Crowding distance

Crowding distance is a popular multi-objective optimization technique that measures the density of non-dominated solutions within a particular region of the objective space. The primary purpose of this technique is to maintain diversity in the population of solutions, preventing them from clustering around a particular region. It works by calculating the average side length of the rectangle formed by the neighboring solutions for each solution. The crowding distance is the sum of the average side lengths in all objective dimensions. In general, the solutions with the highest crowding distances are considered the most diverse and are preferred over other solutions. Typically, the crowding distance of the solutions with the lowest and highest objective function values is assumed to be infinite. A high crowding distance means a better distribution of solutions, while a small crowding distance means that the solution is crowded.

To illustrate the concept of crowding distance, Figure. 1 can be used. It shows the crowding distance of solutions with two objectives. The solutions are sorted in ascending order based on the fitness value of each objective function. The crowding distance of each solution xi is calculated by averaging the side lengths of the rectangles formed by the neighboring solutions $x_{i-1}$ and $x_{i+1}$ adjacent to it. This value represents the crowding distance of solution xi. As can be seen from the figure, the solutions in the dense regions have small crowding distances, indicating that they are crowded, while those in the sparse regions have high crowding distances, indicating that they are more diverse. the peudo-code of the crowding distance is given in figure 2.

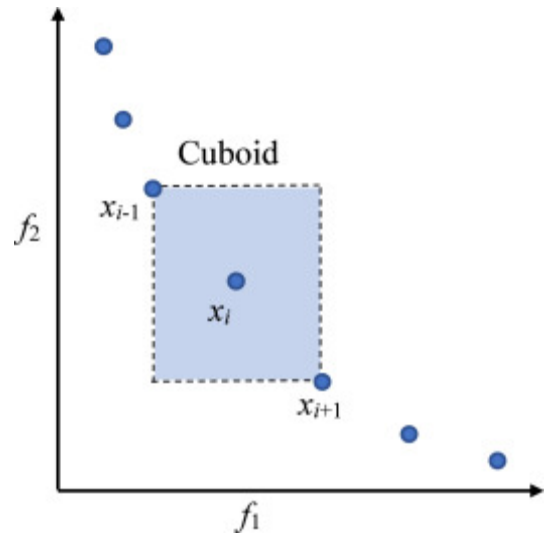

**Figure.1** The crowding distance

**Input:**
A set of non-dominated solutions S with M objectives
The maximum and minimum values of each objective function
**Output:**
Crowding distances for each solution in S
**Algorithm:**
For each solution in S, initialize its crowding distance to 0.
**For each** objective i from 1 to M:
    a. Sort the solutions in S according to their objective i values.
    b. Set the crowding distance of the solutions with the lowest and highest objective i values to infinity.
    c. **For each** solution $x_i$ with objective i value between the lowest and highest:
- Calculate the distance between xi and its two neighboring solutions in objective i dimension.
- Add the average of the two distances to the crowding distance of $x_i$.

**Return** the crowding distances for each solution in S.

**Figure 2.** the crowding distance algorithm

Mathematically, it is computed as follows:

$$d_i(x_i) = \frac{1}{M} \sum_{j=1}^{M} \frac{f_j(x_{i+1}) - f_j(x_{i-1})}{f_j^{max} - f_j^{min}}$$

Where:

- $d_i(x_i)$ is the crowding distance of solution xi
- *M* is the number of objectives
- $f_j(x_i)$ is the value of objective j for solution xi
- $f_j^{max}$ and $f_j^{min}$ are the maximum and minimum values of objective j in the entire population of solutions
- $x_{i+1}$ and $x_{i-1}$ are the neighboring solutions of xi in the objective j dimension
- The crowding distance of the extreme points is set as

$$d_1(x_1) = \infty$$
$$d_N(x_N) = \infty$$

Where:

- $d_1(x_1)$ is the crowding distance of the first sorting element
- $d_N(x_N)$ is the crowding distance of the last sorting element
- N is the number of solutions in the non-dominated set

### 3.2 The modified crowding distances

The traditional method for computing crowding distances involves setting the distance values of the two extreme points to infinity, which can lead to biased results. To address this issue and obtain more accurate results, we have modified the method in two ways. Firstly, we set the distance values of the extreme points to M times the maximum value of the point, where M is the number of objectives. Additionally, we have introduced two artificial points to the solutions: the nadir point, which represents the worst point in all objectives, and the ideal point, which represents the best solution in all objectives. These modifications help to mitigate bias and improve the accuracy of the crowding distance calculation in multi-objective optimization problems. Finally, to accelerate the distance computation, the term $f_j^{max} - f_j^{min}$ is deleted in order to accelerate the distance computation. The modified crowding distance is computed as follows:

$$d_1(x_1) = M * Max(f^{max})$$

$$d_{N+2}(x_{N+2}) = M * Max(f^{max})$$

$$d_i(x_i) = \frac{1}{M} \sum_{j=1}^{M} f_j(x_{i+1}) - f_j(x_{i-1}) \quad i = 2..N+1$$

### 3.3 The deterministic kmeans initialization by crowding distance.

In this work, we propose two new deterministic initialization procedures for clustering algorithms, based on the modified crowding. In the first one, called CKmeans, the initial centroids are selected as follows:

> 1. Compute the crowding distance of all points
> 2. Sort the distance by the ascend order from the high crowded to the less crowded points
> 3. Select the first K points as centroids.

The motivation behind the use of crowding distance in clustering algorithms is that densely populated areas in the objective space often correspond to clusters of similar solutions. By identifying the most crowded points, which typically lie at the core of these clusters, we can better understand the underlying structure of the data and group similar solutions together.

This concept is similar to the density-based clustering algorithm DBSCAN, which identifies clusters based on regions of high density in the data space. In both cases, the idea is to identify regions of the solution space that are densely populated and use these regions as a basis for clustering or grouping similar solutions.

By using crowding distance to identify the most crowded points, we can not only identify the core of each cluster, but also ensure that each cluster is well-separated from other clusters. This can help to improve the quality of the clustering results and make them more interpretable and useful for downstream analysis.

The FCKmeans algorithm is an enhancement of the Ckmeans algorithm, which uses the concept of furthest crowded points (FCPs) to initialize the centroids for the clustering process. The FCPs are selected based on a criterion that considers both the distance between points and their crowding distance. The main advantage of FCKmeans is that it ensures that the densest points in the objective space are far from each other, which can improve the quality of the resulting clusters and prevent the algorithm from converging to suboptimal solutions. The basic outline of the FCKmeans procedure is as follows:

> 1. Compute the crowding distance for each point in the population.
> 2. Sort the points in descending order of their crowding distance.
> 3. Select the first point as the initial centroid.
> 4. For each subsequent point, compute the ratio of its distance to the current centroid and its crowding distance.
> 5. Sort the remaining points in descending order of this ratio.
> 6. Select the point with the highest ratio as the next centroid.
> 7. Repeat steps 4-6 until all centroids have been selected.

By using the FCPs as initial centroids, the FCKmeans algorithm is able to better capture the underlying structure of the data and produce more accurate and interpretable clustering results. Finally, we propose a random initialization version of FCKmeans, called RCKmeans. In this algorithm, the next centroid is selected randomly according to the probability of the ratio distance/crowding.

### 4. Experimental results

In this study, we have implemented the proposed initialization procedures using MATLAB Online 2023 and evaluated their performance on 37 datasets (table.1). These datasets comprise of 17 real datasets and 20 artificial datasets, providing a comprehensive evaluation of the proposed procedures on a diverse range of data types. The characteristics of each dataset are described in detail in Table 1, providing insights into the data distribution, number of clusters, and other relevant information. To assess the effectiveness of the proposed procedures, we have compared them to several popular initialization methods, including Kmeans with random initialization, Kmeans++, MaxMinKmeans with MaxMin initialization, and MinMaxKmeans based on MinMax initialization. Various evaluation metrics were used to evaluate the effectiveness of each initialization procedure for k-means clustering, several metrics can be used. These metrics include:

- Inertia Score (IS): Measures the sum of squared distances between data points and their assigned centroids.
- Rand Index score (RI): Measures the similarity between two clusterings by comparing the number of true positives and true negatives.
- Mutual Information (MI): Measures the amount of information shared between two clusterings.
- Silhouette index score (SI): Measures the quality of clustering by evaluating the distance between data points within clusters and between clusters.
- Davies Bouldin (DB): Measures the average similarity between each cluster and its most similar cluster, while penalizing clusters with high variance.
- Calinski Harabasz (CH): Measures the ratio of between-cluster variance to within-cluster variance.

The results of each initialization procedure can be compared using these metrics. For IS and DB, lower values indicate better performance, while for RI, MI, SI, and CH, higher values indicate better performance. The mean results can be summarized in tables, with the best results indicated in boldface. Finally, the Friedman test is used to rank the different procedures and determine which one performs the best overall.

Based on the results presented in the tables and the Friedman tests, CKmeans and FCkmeans are the top-performing initialization procedures for k-means clustering. They rank first in Inertia, RI, MI, and CH metrics, which indicates that they perform well in terms of cluster quality and similarity to the true clustering. MaxMinKmeans performs better in SI and DB metrics, but is outperformed by CKmeans and FCkmeans in the other metrics. Overall, the superiority of CKmeans and FCkmeans over Kmeans in all tests demonstrates their effectiveness in improving k-means clustering. The high scores in RI and MI metrics indicate that CKmeans and FCkmeans can accurately identify clusters that are closest to the true clustering.

**Table 1:** dataset details

| | dataset | number of samples | number of features | k |
|---|---|---|---|---|
| 1 | iris | 150 | 4 | 3 |
| 2 | ecoli | 336 | 7 | 8 |
| 3 | glass | 214 | 9 | 2 |
| 4 | balance | 625 | 4 | 3 |
| 5 | cancer | 699 | 9 | 2 |
| 6 | ovarian | 216 | 100 | 2 |
| 7 | thyroid | 7200 | 21 | 3 |
| 8 | sonar | 208 | 60 | 2 |
| 9 | chemical_test | 498 | 8 | 40 |
| 10 | ionosphere | 351 | 34 | 2 |
| 11 | data_heart | 267 | 44 | 2 |
| 12 | Zoo | 101 | 16 | 7 |
| 13 | SPECT | 267 | 22 | 2 |
| 14 | COIL20 | 1440 | 1024 | 20 |
| 15 | semeion | 1593 | 265 | 2 |
| 16 | isolet | 1559 | 617 | 26 |
| 17 | house_test | 506 | 13 | 46 |
| 18 | dataset500_2_4 | 500 | 2 | 4 |
| 19 | dataset500_2_5 | 500 | 2 | 5 |
| 20 | dataset500_4_20 | 500 | 4 | 20 |
| 21 | dataset1000_2_4 | 1000 | 2 | 4 |
| 22 | dataset1000_2_5 | 1000 | 2 | 5 |
| 23 | dataset1000_2_10 | 1000 | 2 | 10 |
| 24 | dataset1000_4_3 | 1000 | 4 | 3 |
| 25 | dataset1000_4_20 | 1000 | 4 | 20 |
| 26 | dataset5000_2_4 | 5000 | 2 | 4 |
| 27 | dataset5000_2_10 | 5000 | 2 | 10 |
| 28 | dataset5000_4_3 | 5000 | 4 | 3 |
| 29 | bdataset500_2_5 | 500 | 5 | 2 |
| 30 | bdataset500_2_10 | 500 | 10 | 2 |
| 31 | bdataset1000_2_4 | 1000 | 4 | 2 |
| 32 | bdataset1000_2_10 | 1000 | 10 | 2 |
| 33 | bdataset1000_4_3 | 1000 | 3 | 4 |
| 34 | bdataset1000_4_20 | 1000 | 20 | 4 |
| 35 | bdataset5000_2_4 | 5000 | 4 | 2 |
| 36 | bdataset5000_2_10 | 5000 | 10 | 2 |
| 37 | bdataset5000_4_3 | 5000 | 3 | 4 |

**Table 2.** Inertia mean results

| test | Kmeans | Kmeans++ | CKmeans | FCKmeans | RCKmeans | MaxMinKmeans | MinMaxKmeans |
|---|---|---|---|---|---|---|---|
| 1 | 134.386 | 131.243 | 129.520 | 129.520 | 130.679 | 135.925 | **129.407** |
| 2 | 525.025 | **393.760** | 413.871 | 413.871 | 397.138 | 408.718 | 595.354 |
| 3 | 492.956 | 484.934 | **471.241** | 488.126 | 484.712 | 509.114 | 487.802 |
| 4 | 1007.386 | 1008.237 | 1063.160 | 1063.160 | 1007.436 | 1013.335 | **1006.165** |
| 5 | 1131.552 | 1131.454 | **1128.707** | **1128.707** | 1131.859 | 1132.077 | 1132.193 |
| 6 | 1326.453 | 1326.362 | **1326.219** | 1326.610 | 1326.430 | 1326.440 | 1326.378 |
| 7 | 25584.339 | 25142.727 | 24862.940 | 24862.940 | 24906.776 | 25517.035 | **24694.141** |
| 8 | 1441.403 | 1434.841 | 1429.051 | **1428.408** | 1437.318 | 1439.026 | 1435.718 |
| 9 | 523.113 | 508.142 | 519.097 | 508.049 | **507.635** | 517.971 | 521.898 |
| 10 | 1536.107 | **1536.068** | 1536.158 | 1536.158 | 1536.158 | 1612.035 | 1536.158 |
| 11 | 1412.978 | 1412.766 | **1412.462** | **1412.462** | 1412.782 | 1413.201 | 1413.196 |
| 12 | 223.514 | 219.744 | **219.385** | 219.614 | 224.191 | 224.989 | 225.736 |
| 13 | **1106.897** | 1107.026 | 1108.134 | 1108.134 | 1107.387 | 1107.608 | 1107.222 |
| 14 | 32267.446 | 27680.926 | 27595.535 | 28539.274 | 27750.364 | 30939.197 | **27407.098** |
| 15 | 25154.110 | 25145.762 | **25127.698** | **25127.698** | 25130.753 | 25150.728 | 25177.377 |
| 16 | 30171.132 | **28214.945** | 28385.241 | 28516.284 | 28231.818 | 28960.656 | 28253.812 |
| 17 | 621.886 | 549.936 | 684.460 | 582.240 | **548.313** | 558.125 | 642.557 |
| 18 | 209.369 | 205.525 | **193.817** | **193.817** | 209.509 | 199.700 | 216.965 |
| 19 | 236.268 | 237.418 | 228.487 | **228.350** | 233.817 | 233.850 | 238.297 |
| 20 | 189.739 | 170.236 | 253.506 | **159.286** | 169.860 | 160.484 | 265.006 |
| 21 | 458.980 | 462.823 | 436.035 | 436.035 | 447.496 | 439.830 | **436.006** |
| 22 | 424.380 | 410.340 | 413.042 | **408.260** | 410.408 | 410.391 | 410.223 |
| 23 | 234.124 | 216.877 | 272.051 | **200.722** | 219.350 | 212.997 | 301.863 |
| 24 | 561.528 | 581.104 | **532.151** | **532.151** | 571.304 | 561.532 | 551.746 |
| 25 | 413.970 | 376.098 | 380.238 | **318.284** | 364.895 | 341.827 | 493.730 |
| 26 | 2187.791 | 2116.649 | **1974.367** | **1974.367** | 2134.435 | 2187.791 | 2134.435 |
| 27 | 1129.651 | 1082.348 | 1339.973 | **1033.059** | 1089.522 | 1079.989 | 1178.448 |
| 28 | 3078.825 | 3224.392 | **2933.258** | **2933.258** | 3078.825 | 3661.087 | **2933.258** |
| 29 | 851.028 | **850.980** | 851.034 | 851.034 | 851.025 | 851.028 | 851.034 |
| 30 | **1258.526** | **1258.526** | **1258.526** | **1258.526** | **1258.526** | **1258.526** | **1258.526** |
| 31 | 1466.262 | 1466.387 | 1465.836 | **1465.810** | 1466.287 | 1466.721 | 1465.948 |
| 32 | 2687.197 | 2687.199 | **2687.030** | 2687.119 | 2687.249 | 2687.236 | 2687.331 |
| 33 | 1121.338 | 1120.071 | 1139.300 | 1117.453 | 1119.135 | **1117.180** | 1120.028 |
| 34 | 3651.794 | 3630.347 | 3578.537 | **3578.522** | 3652.550 | 3718.965 | 3712.702 |
| 35 | 8027.733 | 8027.773 | 8028.103 | 8028.103 | 8027.760 | 8027.910 | **8027.685** |
| 36 | 12950.388 | 12950.383 | **12950.265** | 12950.461 | 12950.409 | 12950.415 | 12950.383 |
| 37 | 5591.654 | 5591.402 | **5590.934** | 5593.180 | 5591.647 | 5591.330 | 5591.701 |

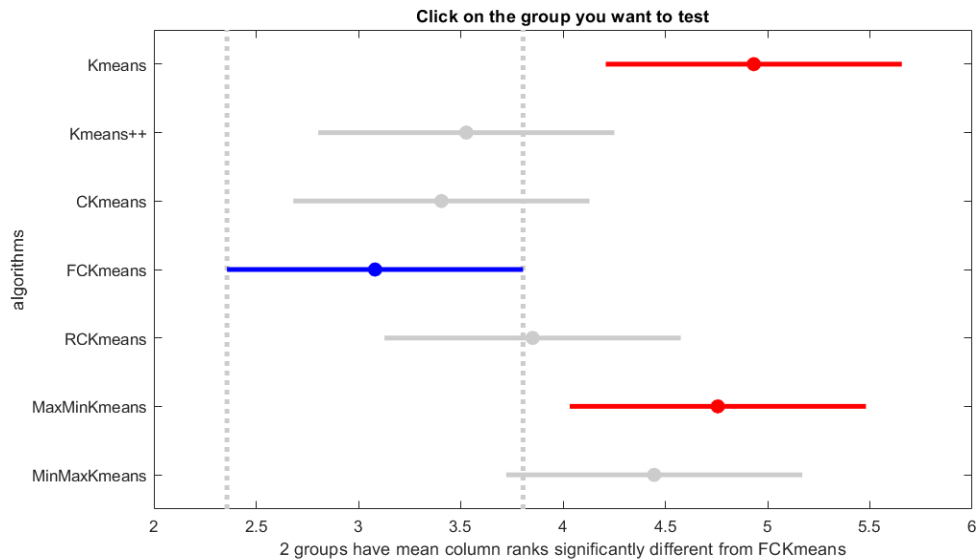

**Figure 3** Friedman test compares Inertia scores

**Table 3.** Inertia RI results

| test | Kmeans | Kmeans++ | CKmeans | FCKmeans | RCKmeans | MaxMinKmeans | MinMaxKmeans |
|---|---|---|---|---|---|---|---|
| 1 | 80.356 | 82.333 | **84.313** | **84.313** | 82.212 | 79.796 | 83.794 |
| 2 | 70.237 | 82.722 | 77.939 | 77.939 | 83.314 | **85.208** | 56.100 |
| 3 | 70.727 | 69.878 | **83.744** | 54.258 | 69.952 | 63.049 | 55.140 |
| 4 | 59.006 | 58.681 | 57.408 | 57.408 | 58.344 | **59.722** | 58.953 |
| 5 | 91.063 | 91.115 | **92.562** | **92.562** | 90.906 | 90.793 | 90.732 |
| 6 | 75.053 | 75.672 | **76.645** | 73.992 | 75.185 | 75.142 | 75.539 |
| 7 | 58.428 | 65.771 | **75.235** | **75.235** | 66.218 | 55.385 | 56.254 |
| 8 | 50.324 | 50.552 | **51.101** | 50.804 | 50.310 | 50.341 | 50.056 |
| 9 | 92.195 | 93.190 | 93.477 | **93.601** | 93.157 | 92.115 | 92.985 |
| 10 | 58.446 | **58.470** | 58.414 | 58.414 | 58.414 | 57.287 | 58.414 |
| 11 | 54.989 | 54.901 | 54.662 | 54.662 | 54.869 | **55.133** | 55.108 |
| 12 | 92.022 | 89.337 | 83.584 | 82.356 | 90.051 | **94.725** | 91.945 |
| 13 | 52.583 | 52.215 | **53.164** | **53.164** | 52.507 | 52.818 | 52.520 |
| 14 | 85.706 | 94.527 | 93.649 | 94.298 | 94.490 | 89.277 | **94.681** |
| 15 | 51.428 | 51.926 | **52.921** | **52.921** | 52.790 | 51.642 | 50.139 |
| 16 | 88.769 | 95.173 | 95.129 | 94.921 | 95.046 | 92.555 | **95.180** |
| 17 | 90.236 | 93.182 | 90.859 | 92.384 | **93.279** | 91.827 | 89.679 |
| 18 | 93.603 | 94.458 | **96.987** | **96.987** | 93.650 | 95.723 | 91.922 |
| 19 | 87.364 | 87.047 | **90.308** | 89.933 | 88.353 | 88.234 | 86.760 |
| 20 | 97.145 | 98.390 | 96.471 | **99.598** | 98.375 | 98.966 | 94.534 |
| 21 | 93.602 | 93.111 | 96.389 | 96.389 | 95.008 | 95.968 | **96.431** |
| 22 | 66.487 | 67.325 | 67.287 | **67.579** | 67.212 | 67.175 | 67.464 |
| 23 | 95.495 | 96.211 | 95.495 | **97.741** | 95.867 | 96.066 | 90.850 |
| 24 | 97.046 | 95.203 | **99.867** | **99.867** | 96.164 | 97.024 | 97.945 |
| 25 | 97.972 | 98.552 | 98.715 | **99.980** | 98.845 | 99.279 | 97.045 |
| 26 | 91.937 | 93.944 | **97.959** | **97.959** | 93.442 | 91.937 | 93.442 |
| 27 | 96.724 | 97.047 | 95.880 | **97.846** | 96.905 | 97.082 | 96.449 |
| 28 | 97.222 | 94.444 | **100.000** | **100.000** | 97.222 | 86.109 | **100.000** |
| 29 | 84.937 | **85.117** | 84.915 | 84.915 | 84.948 | 84.937 | 84.915 |
| 30 | **90.481** | **90.481** | **90.481** | **90.481** | **90.481** | **90.481** | **90.481** |
| 31 | 79.313 | 79.287 | **79.934** | 79.471 | 79.343 | 79.103 | 79.420 |
| 32 | 79.800 | 79.816 | **80.089** | 79.625 | 79.641 | 79.677 | 79.923 |
| 33 | 68.225 | 68.273 | 66.179 | **68.471** | 68.265 | 68.173 | 68.147 |
| 34 | 91.842 | 93.340 | **97.455** | 97.359 | 91.775 | 86.732 | 87.192 |
| 35 | 63.504 | 63.495 | 63.450 | 63.450 | 63.491 | 63.481 | **63.522** |
| 36 | 90.443 | 90.448 | **90.534** | 90.390 | 90.429 | 90.424 | 90.448 |
| 37 | **70.554** | 70.529 | 70.418 | 70.552 | 70.518 | 70.536 | 70.537 |

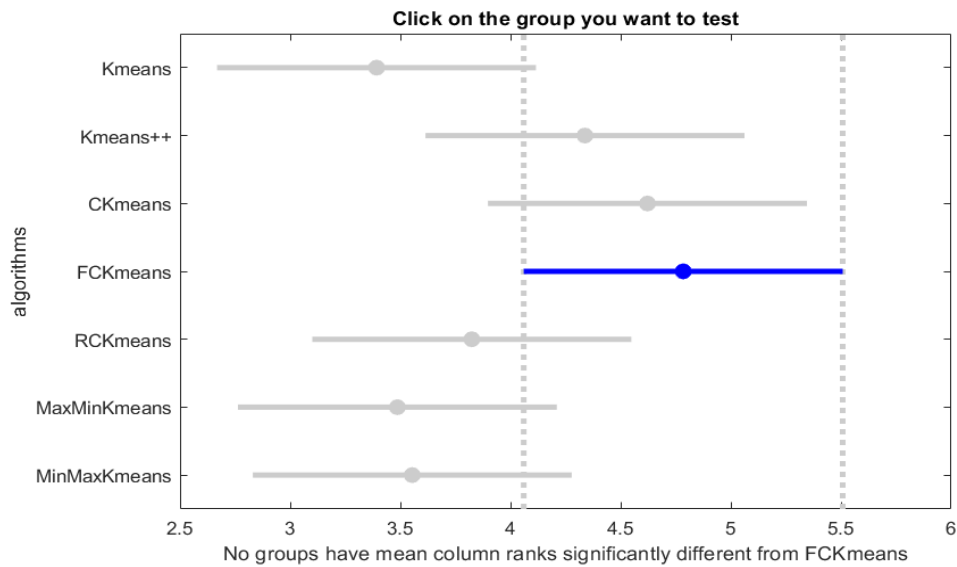

**Figure 4.** Friedman test compares RI scores

**Table 4.** Inertia SI results

| test | Kmeans | Kmeans++ | CKmeans | FCKmeans | RCKmeans | MaxMinKmeans | MinMaxKmeans |
|---|---|---|---|---|---|---|---|
| 1 | **0.653** | 0.650 | 0.644 | 0.644 | 0.649 | 0.649 | 0.646 |
| 2 | 0.587 | 0.415 | 0.334 | 0.334 | 0.361 | 0.491 | **0.601** |
| 3 | 0.657 | 0.573 | 0.589 | 0.426 | 0.495 | **0.721** | 0.424 |
| 4 | 0.282 | 0.280 | 0.139 | 0.139 | 0.279 | 0.269 | **0.287** |
| 5 | 0.722 | 0.722 | 0.718 | 0.718 | 0.721 | 0.723 | **0.723** |
| 6 | 0.652 | 0.651 | 0.650 | **0.654** | 0.652 | 0.652 | 0.652 |
| 7 | 0.325 | **0.371** | 0.316 | 0.316 | 0.277 | 0.278 | 0.261 |
| 8 | 0.346 | 0.315 | 0.216 | 0.224 | 0.293 | 0.375 | **0.392** |
| 9 | 0.352 | 0.329 | 0.282 | 0.297 | 0.309 | **0.379** | 0.317 |
| 10 | 0.374 | 0.374 | 0.374 | 0.374 | 0.374 | **0.434** | 0.374 |
| 11 | 0.665 | 0.663 | 0.659 | 0.659 | 0.660 | **0.668** | 0.667 |
| 12 | 0.503 | 0.461 | 0.445 | 0.448 | 0.429 | **0.557** | 0.456 |
| 13 | 0.305 | 0.305 | **0.322** | **0.322** | 0.309 | 0.314 | 0.307 |
| 14 | 0.282 | 0.313 | **0.314** | 0.224 | 0.284 | 0.294 | 0.297 |
| 15 | 0.104 | 0.106 | **0.110** | **0.110** | 0.109 | 0.104 | 0.098 |
| 16 | **0.167** | 0.141 | 0.120 | 0.122 | 0.136 | 0.161 | 0.133 |
| 17 | 0.374 | 0.364 | 0.273 | 0.380 | 0.349 | **0.421** | 0.358 |
| 18 | 0.774 | 0.780 | **0.805** | **0.805** | 0.763 | 0.793 | 0.754 |
| 19 | 0.624 | 0.614 | 0.659 | **0.664** | 0.630 | 0.635 | 0.615 |
| 20 | 0.706 | 0.722 | 0.603 | 0.738 | 0.703 | **0.753** | 0.666 |
| 21 | 0.746 | 0.743 | **0.769** | **0.769** | 0.754 | 0.765 | 0.769 |
| 22 | 0.672 | 0.671 | 0.672 | 0.670 | 0.673 | **0.674** | 0.669 |
| 23 | 0.735 | 0.750 | 0.680 | 0.748 | 0.736 | **0.758** | 0.752 |
| 24 | 0.837 | 0.828 | **0.852** | **0.852** | 0.833 | 0.837 | 0.842 |
| 25 | 0.711 | 0.733 | 0.790 | **0.837** | 0.712 | 0.777 | 0.682 |
| 26 | 0.734 | 0.747 | **0.774** | **0.774** | 0.754 | 0.734 | 0.744 |
| 27 | 0.776 | **0.783** | 0.718 | 0.771 | 0.762 | 0.772 | 0.780 |
| 28 | 0.822 | 0.811 | **0.833** | **0.833** | 0.822 | 0.778 | **0.833** |
| 29 | 0.458 | 0.458 | **0.458** | **0.458** | 0.458 | 0.458 | **0.458** |
| 30 | **0.419** | **0.419** | **0.419** | **0.419** | **0.419** | **0.419** | **0.419** |
| 31 | 0.506 | 0.506 | 0.504 | 0.505 | 0.506 | **0.507** | 0.506 |
| 32 | 0.333 | 0.333 | 0.333 | **0.334** | 0.333 | 0.333 | 0.332 |
| 33 | 0.374 | 0.377 | 0.334 | **0.383** | 0.378 | 0.382 | 0.375 |
| 34 | 0.244 | 0.237 | 0.232 | 0.233 | 0.240 | **0.265** | 0.262 |
| 35 | 0.401 | 0.401 | **0.402** | **0.402** | 0.401 | 0.401 | 0.401 |
| 36 | 0.373 | 0.373 | 0.373 | **0.373** | 0.373 | 0.373 | 0.373 |
| 37 | 0.376 | 0.376 | 0.377 | 0.370 | **0.377** | 0.377 | 0.375 |

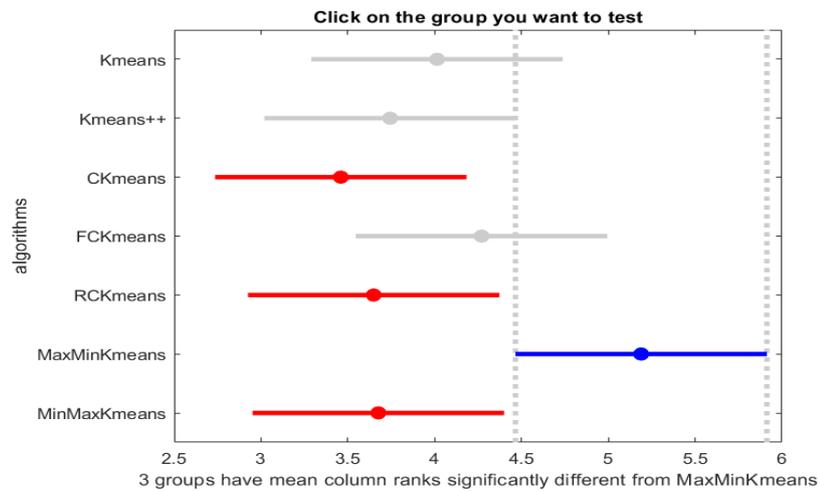

**Figure 5.** Friedman test compares SI scores

**Table 5.** Inertia MI results

| test | Kmeans | Kmeans++ | CKmeans | FCKmeans | RCKmeans | MaxMinKmeans | MinMaxKmeans |
|---|---|---|---|---|---|---|---|
| 1 | 0.688 | 0.710 | **0.726** | **0.726** | 0.707 | 0.678 | 0.724 |
| 2 | 0.658 | 1.041 | 0.964 | 0.964 | **1.044** | 1.009 | 0.437 |
| 3 | 0.105 | 0.113 | **0.255** | 0.000 | 0.108 | 0.010 | 0.001 |
| 4 | 0.124 | 0.113 | 0.079 | 0.079 | 0.109 | **0.141** | 0.118 |
| 5 | 0.455 | 0.456 | **0.482** | **0.482** | 0.453 | 0.451 | 0.450 |
| 6 | 0.329 | 0.335 | **0.345** | 0.318 | 0.330 | 0.330 | 0.334 |
| 7 | **0.012** | 0.011 | 0.001 | 0.001 | 0.008 | 0.011 | 0.009 |
| 8 | 0.011 | 0.012 | **0.014** | 0.011 | 0.011 | 0.013 | 0.010 |
| 9 | 1.310 | 1.388 | 1.385 | **1.412** | 1.373 | 1.282 | 1.339 |
| 10 | 0.084 | **0.085** | 0.084 | 0.084 | 0.084 | 0.063 | 0.084 |
| 11 | 0.033 | 0.034 | **0.035** | **0.035** | 0.034 | 0.033 | 0.033 |
| 12 | 1.215 | 1.229 | 1.108 | 1.111 | **1.229** | 1.228 | 1.219 |
| 13 | 0.021 | 0.018 | **0.024** | **0.024** | 0.020 | 0.022 | 0.021 |
| 14 | 1.477 | 2.031 | **2.124** | 1.982 | 2.017 | 1.645 | 2.039 |
| 15 | 0.010 | 0.012 | **0.018** | **0.018** | 0.017 | 0.011 | 0.004 |
| 16 | 1.606 | **2.111** | 2.066 | 2.078 | 2.095 | 1.923 | 2.100 |
| 17 | 1.183 | 1.378 | 1.289 | 1.347 | **1.388** | 1.269 | 1.210 |
| 18 | 1.200 | 1.215 | **1.263** | **1.263** | 1.199 | 1.239 | 1.168 |
| 19 | 1.082 | 1.078 | **1.151** | 1.140 | 1.108 | 1.102 | 1.068 |
| 20 | 2.656 | 2.781 | 2.568 | **2.912** | 2.779 | 2.847 | 2.414 |
| 21 | 1.173 | 1.162 | 1.229 | 1.229 | 1.202 | 1.222 | **1.231** |
| 22 | 0.007 | 0.008 | 0.008 | 0.008 | 0.007 | 0.007 | **0.008** |
| 23 | 2.002 | 2.041 | 1.968 | **2.110** | 2.028 | 2.041 | 1.768 |
| 24 | 1.044 | 1.014 | **1.092** | **1.092** | 1.030 | 1.043 | 1.058 |
| 25 | 2.751 | 2.821 | 2.830 | **2.986** | 2.855 | 2.906 | 2.620 |
| 26 | 1.152 | 1.197 | **1.287** | **1.287** | 1.186 | 1.152 | 1.186 |
| 27 | 2.083 | 2.102 | 1.995 | **2.136** | 2.096 | 2.103 | 2.074 |
| 28 | 1.052 | 1.006 | **1.099** | **1.099** | 1.052 | 0.867 | **1.099** |
| 29 | 0.457 | **0.459** | 0.457 | 0.457 | 0.457 | 0.457 | 0.457 |
| 30 | **0.526** | 0.526 | 0.526 | 0.526 | 0.526 | 0.526 | 0.526 |
| 31 | 0.375 | 0.375 | **0.380** | 0.375 | 0.375 | 0.373 | 0.376 |
| 32 | 0.394 | 0.394 | **0.397** | 0.392 | 0.392 | 0.392 | 0.395 |
| 33 | 0.236 | 0.235 | 0.196 | **0.236** | 0.234 | 0.229 | 0.231 |
| 34 | 1.121 | 1.155 | **1.252** | 1.248 | 1.119 | 1.006 | 1.015 |
| 35 | 0.175 | 0.175 | 0.175 | 0.175 | 0.175 | 0.175 | **0.175** |
| 36 | 0.523 | 0.523 | **0.524** | 0.522 | 0.522 | 0.522 | 0.523 |
| 37 | 0.313 | 0.313 | 0.309 | 0.309 | 0.312 | **0.314** | 0.312 |

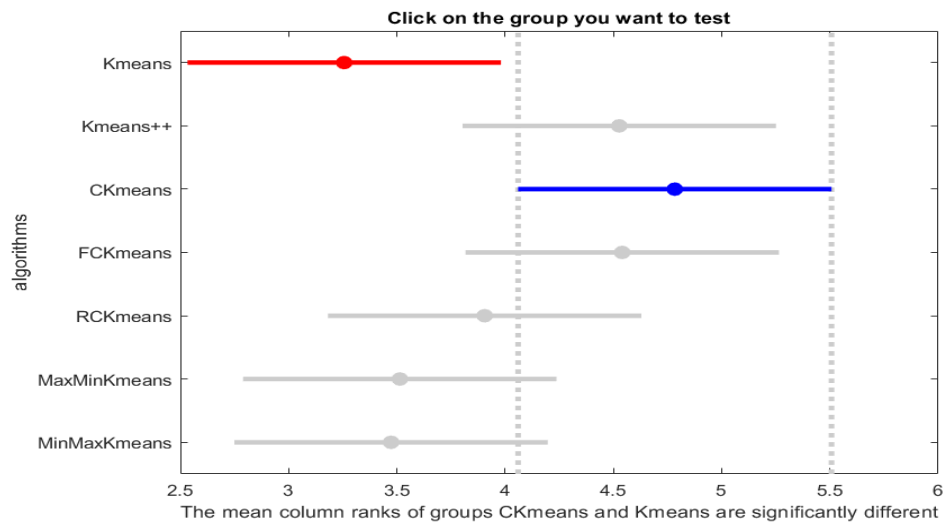

**Figure 6** Friedman test compares MI scores.

**Table 6.** Inertia DB results

| test | Kmeans | Kmeans++ | CKmeans | FCKmeans | RCKmeans | MaxMinKmeans | MinMaxKmeans |
|---|---|---|---|---|---|---|---|
| 1 | **0.802** | 0.818 | 0.830 | 0.830 | 0.830 | 0.812 | 0.832 |
| 2 | 0.596 | 1.113 | 1.301 | 1.301 | 1.147 | 0.970 | **0.591** |
| 3 | 1.200 | 1.310 | 1.358 | 1.670 | 1.314 | **0.839** | 1.687 |
| 4 | **1.708** | 1.718 | 3.186 | 3.186 | 1.711 | 1.773 | 1.723 |
| 5 | 0.824 | 0.824 | **0.823** | **0.823** | 0.824 | 0.824 | 0.824 |
| 6 | 0.826 | 0.826 | 0.828 | **0.824** | 0.826 | 0.826 | 0.826 |
| 7 | 1.924 | 1.767 | **1.219** | **1.219** | 1.674 | 2.043 | 2.195 |
| 8 | 2.220 | 2.285 | 2.452 | 2.438 | 2.254 | 2.167 | **2.086** |
| 9 | 1.173 | 1.275 | 1.395 | 1.352 | 1.273 | **1.105** | 1.307 |
| 10 | 1.682 | 1.681 | 1.682 | 1.682 | 1.682 | **1.355** | 1.682 |
| 11 | 1.509 | 1.513 | 1.527 | 1.527 | 1.516 | **1.501** | 1.503 |
| 12 | 1.187 | 1.307 | **1.127** | 1.212 | 1.331 | 1.142 | 1.386 |
| 13 | 2.203 | 2.201 | **2.174** | **2.174** | 2.195 | 2.189 | 2.201 |
| 14 | **1.525** | 1.824 | 1.812 | 1.892 | 1.792 | 1.554 | 1.837 |
| 15 | 3.957 | 3.865 | **3.671** | **3.671** | 3.702 | 3.918 | 4.212 |
| 16 | **1.554** | 2.564 | 2.686 | 2.768 | 2.605 | 2.099 | 2.617 |
| 17 | 0.917 | 1.156 | 1.206 | 1.175 | 1.156 | **0.901** | 1.026 |
| 18 | 0.615 | 0.594 | **0.534** | **0.534** | 0.614 | 0.565 | 0.653 |
| 19 | 0.808 | 0.815 | 0.717 | **0.703** | 0.774 | 0.778 | 0.841 |
| 20 | 0.847 | 0.816 | 0.836 | 0.745 | 0.802 | **0.690** | 0.840 |
| 21 | 0.626 | 0.634 | 0.577 | 0.577 | 0.601 | 0.585 | **0.577** |
| 22 | 0.806 | 0.782 | 0.804 | **0.754** | 0.788 | 0.784 | 0.776 |
| 23 | 0.709 | 0.682 | 0.740 | **0.638** | 0.688 | 0.676 | 0.653 |
| 24 | 0.553 | 0.586 | **0.502** | **0.502** | 0.570 | 0.553 | 0.536 |
| 25 | 0.857 | 0.791 | 0.687 | **0.533** | 0.755 | 0.676 | 0.883 |
| 26 | 0.703 | 0.665 | **0.589** | **0.589** | 0.674 | 0.703 | 0.674 |
| 27 | 0.631 | 0.618 | 0.647 | **0.602** | 0.633 | 0.633 | 0.625 |
| 28 | 0.574 | 0.595 | **0.553** | **0.553** | 0.574 | 0.659 | **0.553** |
| 29 | 1.448 | **1.448** | 1.448 | 1.448 | 1.448 | 1.448 | 1.448 |
| 30 | **1.600** | **1.600** | **1.600** | **1.600** | **1.600** | **1.600** | **1.600** |
| 31 | 1.313 | 1.313 | 1.314 | 1.313 | 1.313 | **1.313** | 1.313 |
| 32 | 2.206 | 2.206 | 2.207 | 2.206 | **2.205** | 2.206 | 2.206 |
| 33 | 1.223 | 1.213 | 1.339 | **1.193** | 1.207 | 1.196 | 1.214 |
| 34 | 2.410 | 2.327 | **2.032** | 2.032 | 2.395 | 2.786 | 2.766 |
| 35 | 1.796 | 1.796 | **1.794** | **1.794** | 1.796 | 1.795 | 1.796 |
| 36 | 1.790 | 1.790 | **1.790** | 1.790 | 1.790 | 1.790 | 1.790 |
| 37 | 1.203 | 1.203 | **1.202** | 1.204 | 1.204 | 1.203 | 1.204 |

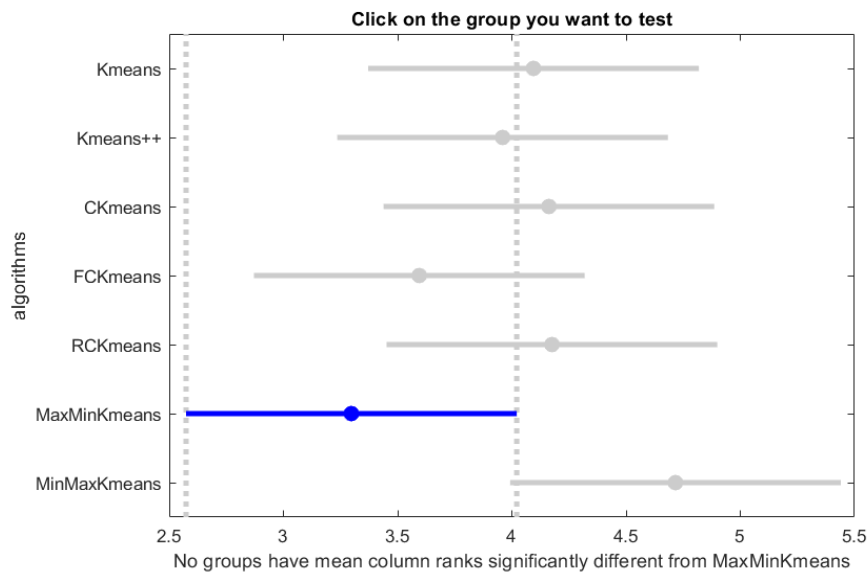

**Figure 7** Friedman test compares DB scores

**Table 7.** Inertia CH results

| test | Kmeans | Kmeans++ | CKmeans | FCKmeans | RCKmeans | MaxMinKmeans | MinMaxKmeans |
|---|---|---|---|---|---|---|---|
| 1 | 215.232 | 229.459 | 236.822 | 236.822 | 232.976 | 209.931 | **237.783** |
| 2 | 78.022 | **140.424** | 81.363 | 81.363 | 125.949 | 140.029 | 45.624 |
| 3 | 40.274 | 43.447 | **55.084** | 42.564 | 43.119 | 31.053 | 42.625 |
| 4 | 135.274 | 134.110 | 90.290 | 90.290 | 135.244 | 126.514 | **136.738** |
| 5 | 868.728 | 868.735 | **868.892** | **868.892** | 868.708 | 868.695 | 868.688 |
| 6 | 264.385 | 264.343 | 264.278 | **264.455** | 264.375 | 264.379 | 264.351 |
| 7 | 433.530 | 410.264 | 393.563 | 393.563 | 402.963 | **441.704** | 413.025 |
| 8 | 31.958 | 33.262 | 31.359 | 31.797 | 33.189 | 33.636 | **35.302** |
| 9 | 61.547 | 65.157 | 57.819 | 61.976 | **65.409** | 64.648 | 59.102 |
| 10 | 95.912 | 95.912 | **95.913** | **95.913** | **95.913** | 71.397 | **95.913** |
| 11 | 81.501 | 81.494 | 81.438 | 81.438 | 81.479 | **81.529** | 81.520 |
| 12 | 28.190 | **28.883** | 26.605 | 26.325 | 27.610 | 28.042 | 27.390 |
| 13 | 52.436 | 52.417 | **52.755** | **52.755** | 52.491 | 52.591 | 52.421 |
| 14 | 68.079 | **103.920** | 99.233 | 89.348 | 103.886 | 78.821 | 103.435 |
| 15 | 92.186 | 93.192 | **95.400** | **95.400** | 95.010 | 92.594 | 89.346 |
| 16 | 36.949 | **51.072** | 49.273 | 47.848 | 50.935 | 45.934 | 50.501 |
| 17 | 64.122 | 85.030 | 41.124 | 63.507 | 85.900 | **85.959** | 57.309 |
| 18 | 1378.217 | 1430.484 | **1586.401** | **1586.401** | 1378.211 | 1508.132 | 1274.183 |
| 19 | 754.581 | 743.836 | **823.574** | 822.754 | 775.681 | 775.791 | 737.793 |
| 20 | 571.652 | 717.030 | 251.804 | 811.489 | 719.513 | **826.930** | 248.715 |
| 21 | 2228.499 | 2192.027 | 2447.281 | 2447.281 | 2337.895 | 2410.828 | **2447.293** |
| 22 | 1969.872 | 2080.731 | 2065.777 | **2096.985** | 2081.555 | 2082.508 | 2083.163 |
| 23 | 3128.892 | 3593.734 | 1844.503 | **4170.932** | 3494.845 | 3685.366 | 1870.616 |
| 24 | 5072.050 | 4837.075 | **5424.541** | **5424.541** | 4954.583 | 5072.040 | 5189.527 |
| 25 | 947.049 | 1156.461 | 1086.584 | **1741.277** | 1245.405 | 1451.484 | 634.855 |
| 26 | 12543.112 | 13385.002 | **15068.784** | **15068.784** | 13174.530 | 12543.112 | 13174.530 |
| 27 | 16463.498 | 17929.471 | 9921.993 | **19627.208** | 17687.353 | 17994.394 | 15172.714 |
| 28 | 18876.527 | 17699.099 | **20053.956** | **20053.956** | 18876.527 | 14166.812 | **20053.956** |
| 29 | 206.453 | 206.449 | **206.453** | **206.453** | **206.453** | **206.453** | **206.453** |
| 30 | **174.915** | **174.915** | **174.915** | **174.915** | **174.915** | **174.915** | **174.915** |
| 31 | 491.319 | 491.322 | **491.332** | 491.311 | 491.322 | 491.317 | 491.309 |
| 32 | 186.960 | 186.956 | **186.976** | 186.973 | 186.958 | 186.959 | 186.892 |
| 33 | 315.727 | 317.722 | 293.203 | **321.855** | 318.531 | 321.251 | 317.658 |
| 34 | 131.038 | 134.335 | 144.332 | **144.338** | 131.168 | 119.451 | 120.395 |
| 35 | 1236.134 | 1236.137 | **1236.138** | **1236.138** | 1236.135 | 1236.136 | 1236.133 |
| 36 | 1405.291 | 1405.291 | **1405.299** | 1405.287 | 1405.290 | 1405.289 | 1405.291 |
| 37 | 1571.127 | 1571.497 | **1571.763** | 1568.493 | 1571.244 | 1571.654 | 1571.006 |

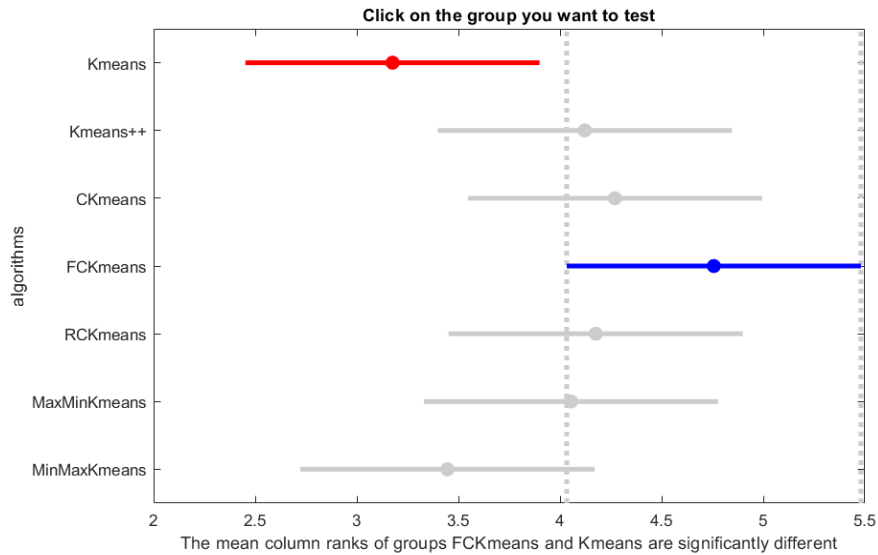

**Figure 8.** Friedman test compares CH scores

## 5. Conclusion

In conclusion, this work proposed two novel initialization procedures for the K-means clustering algorithm called CKmeans and FCKmeans. Both methods utilize a modified crowding distance approach to select the initial centroids. The experimental study showed that the proposed methods outperformed the standard initialization methods of K-means and K-means++ in terms of several metrics, including Inertia, Rand Index, Mutual Information, and Calinski Harabasz. However, the MaxminKmeans method performed better in Silhouette index and Davies Bouldin metrics. The proposed methods can provide a more deterministic and effective initialization procedure for K-means, leading to better clustering results. Further research can investigate the application of the proposed methods in different clustering algorithms and explore their performance in real-world datasets.